%% file: Nakamura_etal_Humanoids2026.tex
\documentclass[letterpaper, 10 pt, conference]{ieeeconf}

\IEEEoverridecommandlockouts
\usepackage{graphicx}
\usepackage{subcaption}
\usepackage{booktabs}
\usepackage{tabularx}
\usepackage{array}
\usepackage{makecell}
\usepackage{amssymb}
\usepackage{cite}
\usepackage{eso-pic}

\title{\LARGE \bf
A Water-Hydraulic, Electronics-Free Robotic Hand for Remote \\Manipulation with Adjustable Transmission-path Compliance
}

\author{%
Yuki Nakamura$^{1,2}$,
Tomoyuki Noda$^{3,1}$,
and Yoshihiro Nakata$^{1}$%
\thanks{This work was partially supported by Chugai Technos Corporation.}%
\thanks{$^{1}$The University of Electro-Communications,
1-5-1 Chofugaoka, Chofu, Tokyo 182-8585, Japan.}%
\thanks{$^{2}$Chugai Technos Corporation,
9-12 Yokogawa-shinmachi, Nishi-ku, Hiroshima 733-0013, Japan.}%
\thanks{$^{3}$Advanced Telecommunications Research Institute International (ATR),
2-2-2 Hikaridai, Seika-cho, Soraku-gun, Kyoto 619-0288, Japan.}%
\thanks{E-mail:
Yuki Nakamura: yuki.nakamura@chugai-tec.co.jp;
Tomoyuki Noda: t\_noda@atr.jp;
Yoshihiro Nakata: ynakata@uec.ac.jp.}%
}
\begin{document}

\maketitle
\thispagestyle{empty}
\pagestyle{empty}

\AddToShipoutPictureFG*{%
  \AtPageLowerLeft{%
    \put(\LenToUnit{0.75in},\LenToUnit{0.30in}){%
      \parbox[b]{7in}{%
        \normalfont\fontsize{8}{9}\selectfont
        \raggedright
        \copyright~2026 IEEE.
        Personal use of this material is permitted.
        Permission from IEEE must be obtained for all
        other uses, in any current or future media,
        including reprinting/republishing this material
        for advertising or promotional purposes,
        creating new collective works, for resale or
        redistribution to servers or lists, or reuse of
        any copyrighted component of this work in
        other works.
      }%
    }%
  }%
}

\begin{abstract}
Remote manipulation in hazardous environments requires both firm grasping and compliant contact, creating a tradeoff between forceful object handling and accommodation of imposed motion. This paper presents a human-scale opposing-finger water-hydraulic hand using hydrostatic transmission (HST): remotely generated pressure drives hand-side hydraulic actuators through long, small-diameter water-filled tubes, without hand-side electronics. Air injection and removal upstream of water injection, with a tube interval between the injection points, tune the effective transmission-path compliance within the same architecture. Single-finger force/pressure--position responses were evaluated through 20 and 50~m main sections under No air, Low air, and High air. For both lengths, High air reduced the pressure gradient from the No-air range of 0.0272--0.0326~MPa/mm to 0.00377--0.0179~MPa/mm. The measured maximum fingertip force was 144~N at 7.5~MPa, beyond the intended design operating points up to 5~MPa. Qualitative three-finger demonstrations through a 50~m tube under No air included bolt pickup, tool grasping, and toolbox-handle holding. Under High air, a single finger accommodated an externally displaced egg while maintaining contact without observed breakage. These observations illustrate firm grasping and compliant contact within one remote actuation architecture.

\end{abstract}

\section{INTRODUCTION}
Hazardous environments require robots that can perform physical work while
operators remain at a safe location. The Fukushima Daiichi response
demonstrated this need for remote reconnaissance and debris handling
\cite{Kawatsuma2012FukushimaRobots,Nagatani2013FukushimaRobots}. Mobile
manipulators must therefore identify, grasp, and retrieve unknown objects while
retaining human expertise in the control loop
\cite{Roennau2022HazardousObjectGrasping}.

Remote manipulation of unknown objects should preserve familiar hand-centered strategies. Existing electric, hydraulic, tendon-driven, and fluidic hands have established the feasibility of dexterous, forceful, remote, and electronics-free manipulation \cite{Jacobsen1984UtahMitHand,Miyamoto2011HydraulicHand,Mori2010PowerRobotHand,Azami2023HydraulicSoftHand,Ishibashi2025DualArmCabbage,Zhou2025DexCo,Endo2019SuperDragon,Drotman2021ElectronicsFree,Nakamura2025RemoteWaterHydraulic,Wang2022WaterHydraulicHand,Ishibashi2023WaterPumpSoftHand}. Limitations include difficult environmental sealing, oil leakage concerns, and friction or elongation in long tendons, motivating an approximately human-scale, multi-degree-of-freedom end effector with sufficient force and resistance to water, impact, debris, and contamination.

\begin{figure}[t]
  \centering
    \begin{minipage}[t]{0.7\hsize}
        \centering
        \includegraphics[width=\linewidth]{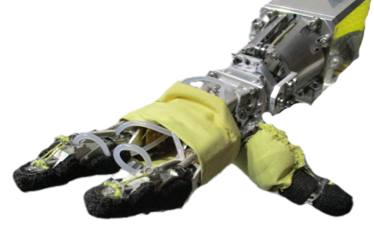}
    \end{minipage} 

  \caption{Proposed electronics-free water-hydraulic robotic hand.}
  \label{fig:hand_photo}
\end{figure}
Rigid and compliant hands serve complementary tasks: rigidity supports forceful grasping, tool holding, valve turning, and pushing, whereas compliance accommodates shape errors and imposed motion during contact with uncertain objects or the environment. Remote manipulation requires both capabilities, motivating adjustment of effective transmission-path compliance to address this tradeoff. The proposed architecture varies the compressed-air volume in the water-pressure path without placing sensors or electronics in the hand.


\begin{table*}[!t]
\caption{Comparison of fluid-powered robotic hands in terms of
remote actuation, electronics-free hand modules, and fluid-path
compliance adjustment.}
\label{tab:hand_comparison}
\centering
\scriptsize
\setlength{\tabcolsep}{3.2pt}
\renewcommand{\arraystretch}{1.06}

\newcommand{\tablehead}[1]{%
  \parbox[c][2.8\baselineskip][c]{\linewidth}{%
    \centering\strut #1\strut}}

\newcommand{\criteriondetail}[2]{%
  \parbox[t]{\linewidth}{%
    \centering
    \strut #1\\[-0.4ex]
    {\tiny #2}}}

\newcommand{\criterionmark}[1]{%
  \parbox[t]{\linewidth}{%
    \centering
    \strut #1}}

\begin{tabularx}{\textwidth}{
@{}
>{\raggedright\arraybackslash}p{0.145\textwidth}
>{\raggedright\arraybackslash}X
!{\hspace{3pt}}
>{\raggedright\arraybackslash}p{0.20\textwidth}
>{\centering\arraybackslash}p{0.085\textwidth}
>{\centering\arraybackslash}p{0.115\textwidth}
>{\centering\arraybackslash}p{0.125\textwidth}
@{}}
\toprule

\tablehead{Reference} &
\tablehead{Hand / grasp configuration} &
\tablehead{Actuation principle} &
\tablehead{Remote\\actuation} &
\tablehead{Electronics-free\\hand} &
\tablehead{Fluid-path compliance\\adjustment} \\

\midrule

Jacobsen et al.~\cite{Jacobsen1984UtahMitHand} &
\makecell[l]{3 fingers + 1 opposable thumb;\\
anthropomorphic opposition;\\
16 active DOFs (4/digit)} &
\makecell[l]{Pneumatic cylinders;\\
tendon transmission} &
$\checkmark$ &
$\times$ &
$\times$ \\

\addlinespace[5pt]

Azami et al.~\cite{Azami2023HydraulicSoftHand} &
\makecell[l]{4 fingers; radial wrapping grasp;\\
common-pressure coupled bending} &
\makecell[l]{Oil-hydraulic\\
fiber-constrained soft actuators} &
$\times$ &
$\times$ &
$\times$ \\

\addlinespace[5pt]

Ishibashi et al.~\cite{Ishibashi2025DualArmCabbage} &
\makecell[l]{3 fingers; radial wrapping grasp;\\
common-pressure coupled bending} &
\makecell[l]{Oil-hydraulic\\
fiber-constrained soft actuators} &
$\times$ &
$\times$ &
$\times$ \\

\addlinespace[5pt]

Zhou et al.~\cite{Zhou2025DexCo} &
\makecell[l]{2 opposing fingers;\\
6 finger DOFs + 1 palm DOF} &
\makecell[l]{Water-hydraulic origami actuators;\\
pneumatic palm} &
$\checkmark$ &
$\times$ &
$\times$ \\

\addlinespace[5pt]

Wang et al.~\cite{Wang2022WaterHydraulicHand} &
\makecell[l]{5 fingers + secondary thumb;\\
bidirectional opposition;\\
4 inputs/digit + 3 palm actuators} &
\makecell[l]{Water-hydraulic\\
bellows-type soft actuators} &
$\checkmark$ &
$\mathrm{NR}$ &
$\times$ \\

\addlinespace[5pt]

Ishibashi et al.~\cite{Ishibashi2023WaterPumpSoftHand} &
\makecell[l]{4 fingers; radial wrapping grasp;\\
common-pressure coupled bending} &
\makecell[l]{Water-hydraulic\\
fiber-constrained soft actuators} &
$\times$ &
$\times$ &
$\times$ \\

\addlinespace[5pt]

Mori et al.~\cite{Mori2010PowerRobotHand} &
\makecell[l]{3 fingers; enveloping grasp;\\
coupled 5-link bending} &
\makecell[l]{Water-hydraulic\\
McKibben muscles} &
$\checkmark$ &
$\checkmark$ &
$\times$ \\

\addlinespace[5pt]

Miyamoto et al.~\cite{Miyamoto2011HydraulicHand} &
\makecell[l]{3-finger surgical retractor;\\
3 DOFs total; 3 fingertip joints\\
share 1 hydraulic DOF} &
\makecell[l]{Water-hydraulic fingertip cylinders;\\
compliance by clipping\\
elastic silicone tube} &
$\checkmark$ &
$\checkmark$ &
$\checkmark$ \\

\midrule

\textbf{This work} &
\makecell[l]{\textbf{3 fingers; 1-versus-2 opposing}\\
\textbf{grasp configuration;}\\
\textbf{7 actuated DOFs (3/3/1)}} &
\makecell[l]{\textbf{Water-hydraulic}\\
\textbf{piston--cylinder joints}} &
\criteriondetail{\textbf{$\checkmark$}}
{\textbf{20 and 50\,m}} &
\criterionmark{\textbf{$\checkmark$}} &
\criteriondetail{\textbf{$\checkmark$}}
{\textbf{Air-volume tuning}} \\

\bottomrule
\end{tabularx}

\vspace{0.8mm}
\begin{minipage}{\textwidth}
\fontsize{6.7}{7.8}\selectfont
\setlength{\parindent}{0pt}
\setlength{\parskip}{0pt}

\textit{Notes:}
$\checkmark$, $\times$, and $\mathrm{NR}$ denote criterion satisfied,
criterion not satisfied, and insufficient information for determination,
respectively.
``Remote actuation'' requires the actuation source to be physically
separated from the hand.
``Electronics-free hand'' excludes motors, valves, sensors, controllers,
and electronic circuits integrated into the hand.
``Fluid-path compliance adjustment'' requires modification of a
fluid-path parameter; inherent softness, pressure-dependent stiffness
without fluid-path modification, and antagonist actuation alone do not
satisfy this criterion.
For this work, single-finger force evaluations used 20~m and 50~m tubes,
whereas the three-finger task demonstrations used a 50~m tube.

\end{minipage}
\end{table*}

The proposed human-scale water-hydraulic hand (Fig.~\ref{fig:hand_photo}) has opposing fingers and no active electronics or sensors at the hand. Its hydrostatic transmission (HST) conveys remotely generated pressure through long, small-diameter water-filled tubes to hand-side hydraulic actuators. Air injection and removal adjust the effective transmission-path response. Main transmission sections extend up to 50~m, with intended design operating points up to 5~MPa. Opposition supports firm grasping and pinching, while air adjustment supports accommodation of externally imposed contact motion.

The contributions of this study are as follows.
First, the hand integrates remote high-pressure transmission through thin water-filled tubes with air injection/removal to tune effective transmission-path compliance, quantitatively characterized by single-finger force/pressure--position tests through 20~m and 50~m main sections.
Second, a human-scale opposing-finger hand integrates seven actuated degrees of freedom (DOFs) for firm grasping and pinching. The finger base joint provides $101^\circ$ flexion and $21^\circ$ extension, and the measured maximum fingertip force is 144~N at 7.5~MPa.
Third, qualitative demonstrations illustrate the complementary capabilities of the integrated architecture: a three-finger hand performs firm bolt, tool, and toolbox-handle tasks through a 50~m tube under No air, while a single finger under High air accommodates external displacement of an egg while maintaining contact.

\section{RELATED WORK}
Table~\ref{tab:hand_comparison} compares fluid-powered robotic hands in terms of remote actuation, electronics-free hand modules, and fluid-path compliance adjustment. This work evaluates single-finger responses through 20~m and 50~m water tubes and demonstrates three-finger tasks through a 50~m tube. The following discussion provides the technical context.

\subsection{Pneumatic and Tendon-Driven Dexterous Hands}
The UTAH/M.I.T. hand pioneered a multi-DOF anthropomorphic design driven by remotely located pneumatic actuators through tendon transmission \cite{Jacobsen1984UtahMitHand}. For remote actuation, tendon-driven mechanisms, such as the Super Dragon manipulator, have demonstrated long-distance mechanical transmission that separates heavy actuators from the distal mechanism \cite{Endo2019SuperDragon}. These systems provide precise, multi-DOF dexterous control, even though long tendons are susceptible to introducing friction, elongation, and routing complexities into constrained spaces.

\subsection{Hydraulic and Soft Hydraulic Robotic Hands}
Compact hydraulic systems, hydraulic McKibben actuators, and soft hydraulic hands offer distinct configurations for generating large forces and adapting to object shapes \cite{Miyamoto2011HydraulicHand,Mori2010PowerRobotHand,Azami2023HydraulicSoftHand,Ishibashi2025DualArmCabbage,Zhou2025DexCo}. Collectively, these studies established fluid power as an effective means of achieving strong, impact-resistant, and compliant grasping that conforms to uncertain object geometries.

\subsection{Water-Hydraulic and Underwater Robotic Hands}
Remote water-hydraulic drives and water-powered hands provide a specialized fluidic context in which water serves as the working fluid \cite{Nakamura2025RemoteWaterHydraulic,Wang2022WaterHydraulicHand,Ishibashi2023WaterPumpSoftHand}. The primary advantages of water hydraulics include low fluid viscosity, which minimizes viscous losses over long distances, low environmental impact, and low risk of secondary contamination in the event of fluid leakage.

\subsection{Electronics-Free Fluidic Robots and Compliance}
Electronics-free pneumatic circuits generate useful robot behaviors through physical fluid dynamics without local electronic components \cite{Drotman2021ElectronicsFree}. The reported soft pneumatic circuits operate at lower pressures than the intended water-hydraulic design points here. Gas compressibility makes high-stiffness transmission more demanding than with nominally incompressible water, while output force depends on operating pressure and actuator area; this is not a universal pneumatic pressure limit. Accumulator- and piston--linkage-based variable transmissions provide other physical approaches to compliance adjustment \cite{park2023stiffness,gan2020modeling}.

Remote actuation, an electronics-free hand, and fluid-path compliance adjustment already coexist in Miyamoto et al.~\cite{Miyamoto2011HydraulicHand}. Their three-finger surgical retractor has three DOFs in total, including one hydraulic DOF shared by three fingertip joints, and adjusts compliance by clipping an elastic silicone tube. The reported configuration is not an opposing grasp, and firm opposing grasping was not assessed. Such elastic-section tuning requires operation within the section's pressure capability; no quantified tube pressure rating is established here. The present contribution is the integration of long-distance, high-pressure water-hydraulic transmission, an opposing seven-actuated-DOF hand, and upstream air injection/removal for path tuning, rather than compliance tuning alone. Quantitative single-finger tests through 20~m and 50~m main sections and qualitative three-finger tasks through a 50~m tube evaluate this integration, not matched pressure-capacity or grasp-performance superiority over the retractor. The air-setting valves and disconnection are operated manually.

\section{METHODS}
\subsection{System Concept and Design}
The proposed system separates the hand-side mechanism from the control module. In this HST architecture, remotely generated pressure is transmitted through thin water-filled tubes to the hand-side hydraulic actuators. The control module can be installed in a safer or more protected location.

\begin{figure}[t]
  \centering
    \begin{minipage}[t]{1.0\hsize}
        \centering
        \includegraphics[width=\linewidth]{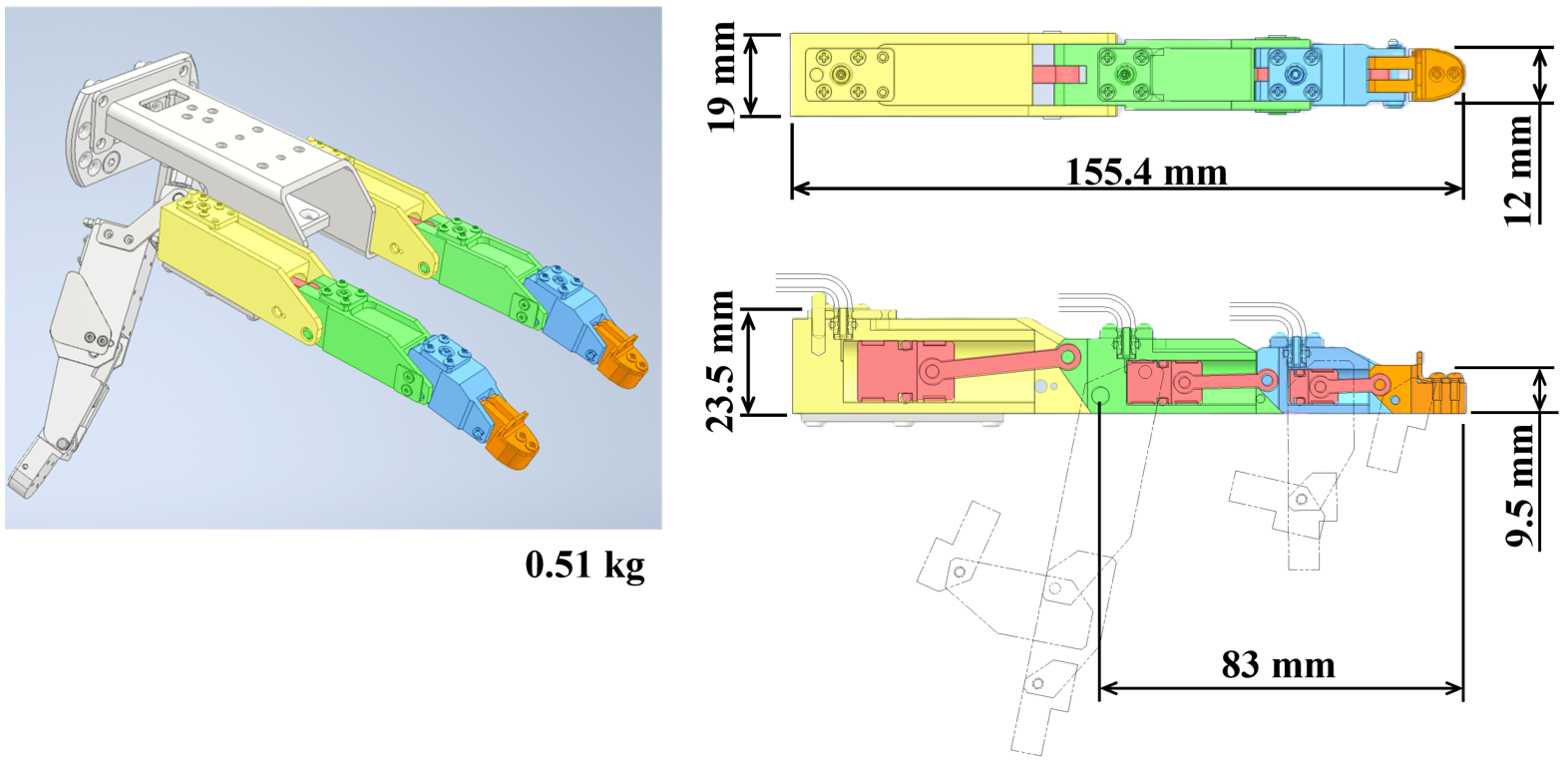}
    \end{minipage} 
  \caption{External view of the hand and structural diagram of the finger.}
  \label{fig:finger_diagram}
\end{figure}

Fig.~\ref{fig:finger_diagram} shows an external view of the proposed hand and a structural diagram of the index finger. This paper evaluates the compliance adjustment at the base joint of this finger.
The index finger can move $101^\circ$ of flexion and $21^\circ$ of extension at the base joint from the position shown in Fig.~\ref{fig:finger_diagram}.

Regarding the base joint, the design dictates that applying a water pressure of 5 MPa generates a torque of 9.2 N·m, resulting in a force of 110 N at the fingertip, 83 mm from the bending axis.

\begin{figure}[t]
  \centering
    \begin{minipage}[t]{1.0\hsize}
        \centering
        \includegraphics[width=\linewidth]{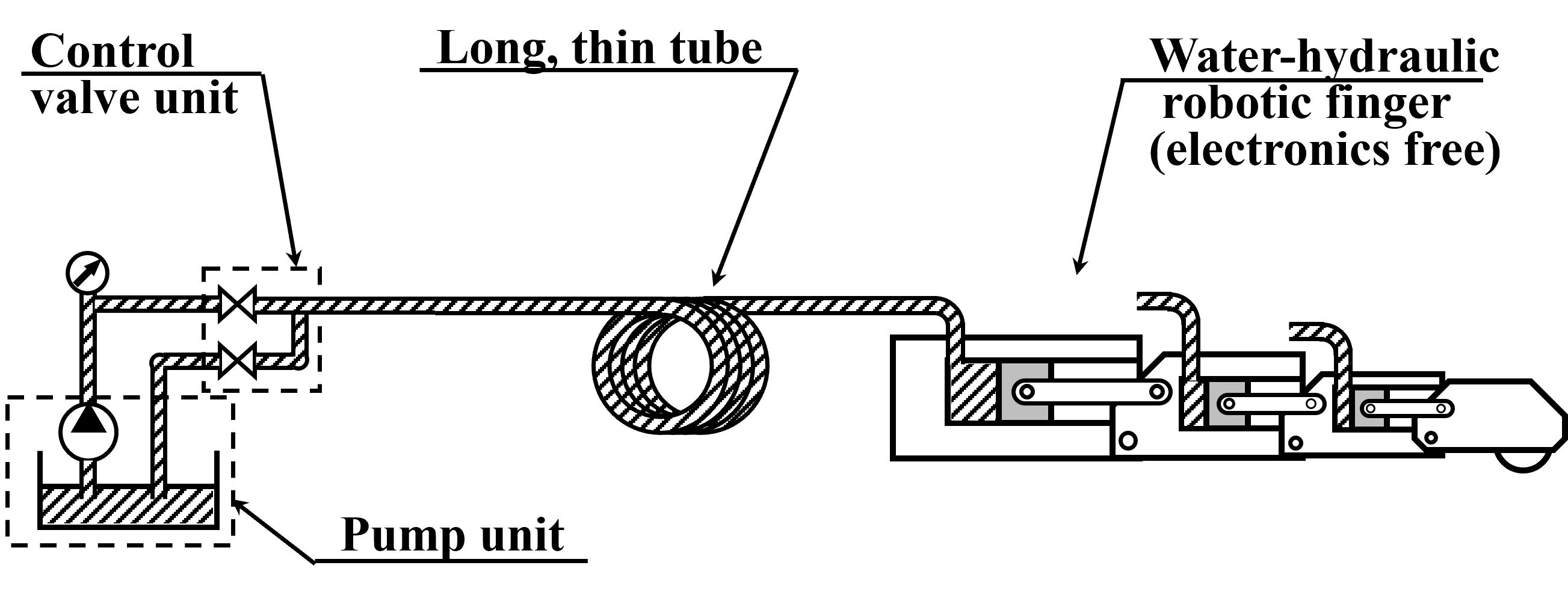}
    \end{minipage} 
  \caption{Configuration of a water-hydraulic finger. A hydraulic-pressure supply tube is connected to each of the three joints, although only one tube is shown. Hydraulic pressure flexes the finger, and a return spring extends it; the spring is not shown.}
  \label{fig:finger_configuration}
\end{figure}

Each finger is treated as a repeatable unit that can be arranged into a multifinger hand. The single-finger configuration illustrated in Fig.~\ref{fig:finger_configuration} shows the joint and link layouts, hydraulic paths, control-side connections, and finger-side mechanisms used in the unit.
The same unit mechanism is intended to be scaled from a single-finger experiment to a human-scale hand, and the number of tubes and pressure channels determines the available DOFs.

\subsection{Water-Hydraulic Finger with Pneumatic Compliance} \label{circuit_air_injection}
In an ideal water-hydraulic transmission line containing no gas, the fluid path is nearly incompressible, and the finger behaves rigidly. Such rigidity can be useful for forceful grasping or tool holding. However, contact tasks require compliance to avoid excessive impact forces and conform to uncertain object geometries.

\begin{figure}[t]
  \centering
    \begin{minipage}[t]{0.8\hsize}
        \centering
        \includegraphics[width=\linewidth]{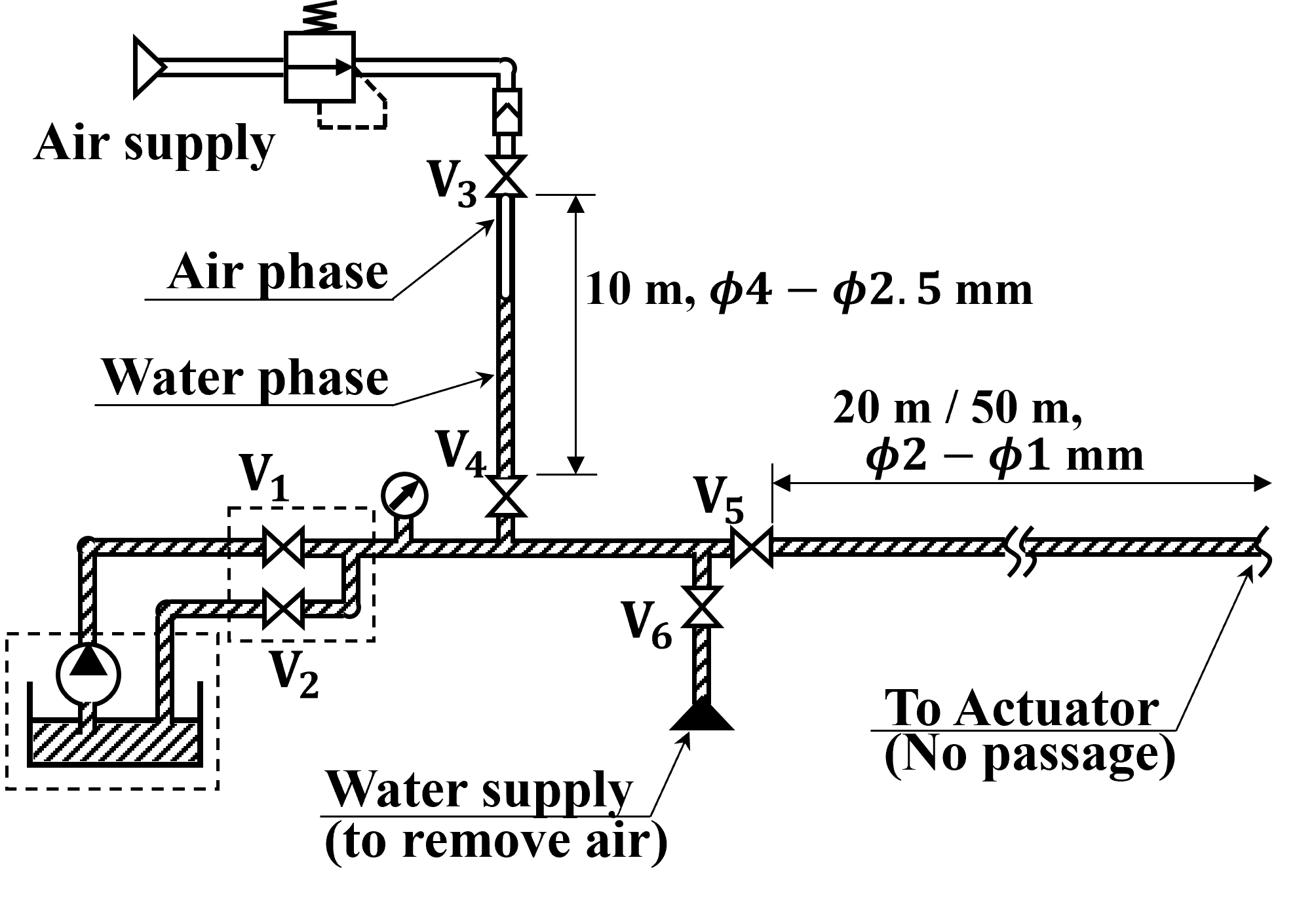}
    \end{minipage} 
  \caption{Implementation of compliance adjustments}
  \label{fig:air_compliance_principle}
\end{figure}

To adjust compliance, air is introduced into the hydraulic drive path upstream of the water injection, separated by a tube interval (Fig.~\ref{fig:air_compliance_principle}).
This layout modifies the effective transmission-path response and is intended to reduce the risk of air reaching the actuator.
During air--water co-injection, valve $V_3$ is closed and $V_4$ is opened.
This design avoids accumulators or extra pistons, though piston friction still affects the measured fingertip force.
To reduce compliance, valve $V_6$ is opened to flush air toward $V_3$ with water.
Here, the fitting above $V_3$ was disconnected for discharge, though a solenoid valve could automate this process.
While these tests confirmed the intended pressure-path adjustment and recovery, air exclusion at the actuator was a design rationale rather than directly measured.

\section{EXPERIMENTS}
\subsection{Stiffness Evaluation}

\begin{figure}[t]
  \centering
    \begin{minipage}[t]{1.0\hsize}
        \centering
        \includegraphics[width=0.9\linewidth]{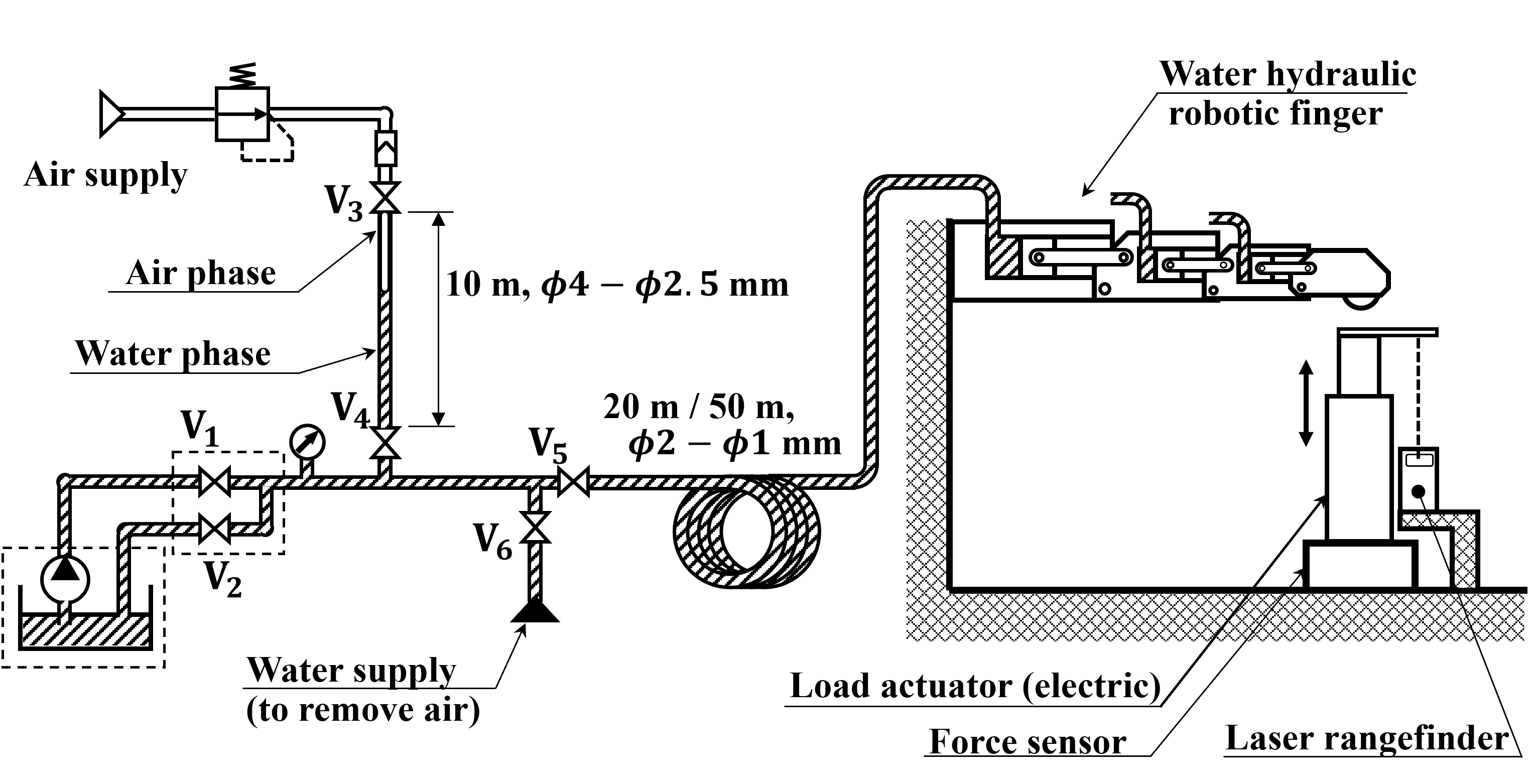}
        \subcaption{For compliance adjustment}
    \end{minipage}
    \\[1.5mm]
    \begin{minipage}[t]{0.9\hsize}
        \centering
        \includegraphics[width=\linewidth]{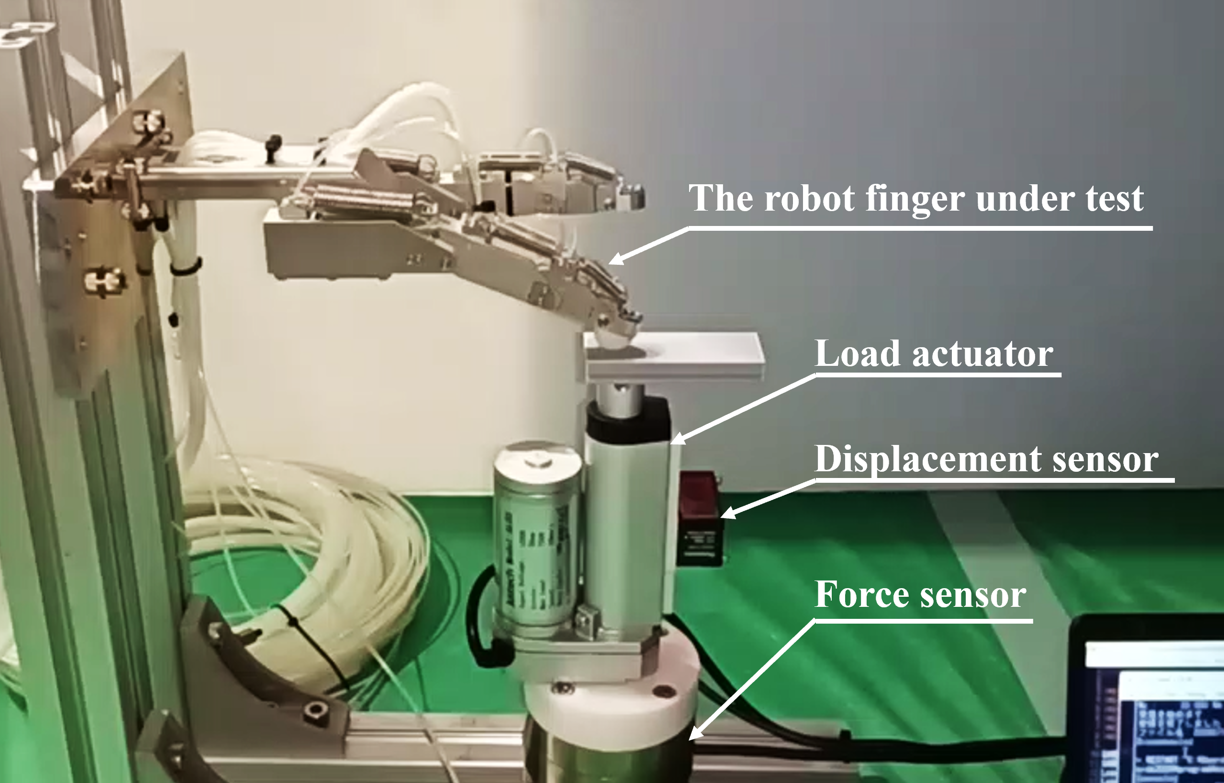}
        \subcaption{Robot finger and load equipment}
    \end{minipage}
  \caption{Experimental setup.}
  \label{fig:experimental_setup}
\end{figure}

The following evaluations were conducted using the proposed system:
\begin{itemize}
    \item Verification of compliance adjustment
    \item Recovery after air removal
    \item Comparison of coupled transmission configurations
\end{itemize}
The experimental setup was designed to evaluate the stiffness of the robotic finger base joint.
As shown in Fig.~\ref{fig:experimental_setup}, the apparatus consists of a water-hydraulic finger, remote pressure-actuation hardware, slender water tubes, a load actuator for imposing displacement, and devices for measuring the fingertip force and displacement.
The load actuator pushes the fingertip upward while force and position are recorded.
At this time, the base joint of the robot finger moves, and the fingertip lifts up.
The two distal joints cannot move further upward because they would have reached their mechanical limits.
The specifications for each device are listed in Table~\ref{table_equipment}.
Figure~\ref{fig:experimental_setup}(a) distinguishes the separate 10~m air-setting section (OD 4~mm, ID 2.5~mm) from the downstream main transmission section (OD 2~mm, ID 1~mm).
The 20~m and 50~m labels refer to the latter section, not a measured total circuit length.
Changing this section changes both downstream tube length and injection-region-to-actuator separation.


We define the condition-specific empirical force slope $k_F$ (N/mm) and
pressure--position gradient $g_P$ (MPa/mm) by the regressions
\begin{equation}
 F=k_Fx+b_F, \qquad P=g_Px+b_P,
 \label{eq:stiffness_definition}
\end{equation}
where $F$ is fingertip force, $P$ is tube pressure, and $x$ is load-actuator displacement. Both force--position and pressure--position responses are nonlinear and hysteretic. Each regression uses ordinary least squares over all valid samples of the full recorded trajectory, combining loading and unloading, to give a representative slope over the full measured displacement interval. The two directions are plotted separately to expose hysteresis; the reported slope is not an arithmetic average of separately fitted branch slopes. Thus, $k_F$ is an empirical descriptor of the complete measured fingertip response, rather than an identified elastic stiffness, and $g_P$ describes the effective pressure-path response. Neither slope is a local tangent at an operating point. The coefficient of determination is calculated as

\begin{equation}
 R_Y^2 = 1-\frac{\sum_i(Y_i-\widehat{Y}_i)^2}
 {\sum_i(Y_i-\overline{Y})^2}, \qquad Y\in\{F,P\},
 \label{eq:r2_definition}
\end{equation}
where $\widehat{Y}_i$ is the fitted response and $\overline{Y}$ is its sample
mean. The inverse pressure descriptor is
\begin{equation}
 C_P = \frac{1}{g_P},
 \label{eq:compliance_definition}
\end{equation}
with units of mm/MPa. It characterizes the effective pressure-path position response, to which tube expansion and compression of introduced air contribute. The pressure fits provide the primary quantitative summary of transmission-path response.

\begin{table}[tbp]
\centering
\caption{Experimental equipment}
\label{table_equipment}
\scriptsize
\setlength{\tabcolsep}{2.5pt}
\renewcommand{\arraystretch}{1.05}
\begin{tabularx}{\columnwidth}{@{}>{\raggedright\arraybackslash}p{0.27\columnwidth} >{\raggedright\arraybackslash}X >{\raggedright\arraybackslash}p{0.20\columnwidth}@{}}
\toprule
\textbf{Equipment} & \textbf{Model or specification} & \textbf{Manufacturer} \\
\midrule
Tube (long section) & AS1, OD 2 mm, ID 1 mm, 20 or 50~m & Junkosha Inc. \\
Tube (pneumatic section) & AS1, OD 4 mm, ID 2.5 mm, 10~m & Junkosha Inc. \\
Load actuator & Maximum 750~N, 50~mm stroke & Pudibei \\
Laser rangefinder & HG-C1100 & Panasonic \\
Force sensor & PFS080YA501R6 & Leptrino \\
Pressure sensor & SPRV-010MP-02-V & Sentec \\
\bottomrule
\end{tabularx}
\end{table}

\subsubsection{Verification of compliance adjustments}
Fingertip force and tube-pressure responses to displacement were measured for nominal air settings and initial hydrostatic pressures.
First, using the circuit described in Section \ref{circuit_air_injection}, air is injected into the  water hydraulic circuit under atmospheric pressure.
With valve $V_2$ open, air valve $V_3$ was slightly opened and then closed when the leading air bubbles visually reached 2~m or 4~m along the larger-bore air-setting section.
Based on its inner diameter, Low air (2~m) and High air (4~m) correspond to nominal calculated reference volumes of approximately 9.8~mL and 19.6~mL at atmospheric pressure, respectively, not in-situ volume measurements.
These settings change nominal air amount and interface position together; neither effect is isolated, and the air--water distribution and mixing during pressurization were not measured.


Next, the load was applied and measured: Valve $V_1$ was opened, $V_2$ was closed, and the pump pressure was adjusted to supply initial pressure to the robot finger, bringing the fingertip into contact with the load actuator tip.
Valve $V_1$ was then closed to seal the system.
The load actuator was driven at a fixed electrical setting ($12~\mathrm{V}$, $20\%$ PWM duty cycle), not a measured constant speed, to move the fingertip up and down through one full cycle between $10~\mathrm{mm}$ and $50~\mathrm{mm}$.
Position, force, and internal tube pressure were recorded at $20~\mathrm{Hz}$ to evaluate their relationship under each condition.
Acquisition proceeded sequentially from No air to Low air to High air, with initial pressures of 0.5, 1.0, and 1.5~MPa in that order for each air condition; it was not randomized.

\subsubsection{Recovery observation of compliance adjustments}
To observe recovery after air removal with the 20~m main transmission section, two full cycles of air injection and removal followed the sequence:
No air $\rightarrow$ Low air (2~m) $\rightarrow$ High air (4~m) $\rightarrow$ Air removal $\rightarrow$ No air
$\rightarrow$ Low air (2~m) $\rightarrow$ High air (4~m) $\rightarrow$ Air removal $\rightarrow$ No air.

Air was removed to restore the fully water-hydraulic state using the following procedure:
First, the hydraulic system was depressurized to atmospheric pressure, and valve $V_5$ was closed. Next, valve $V_3$ was opened, its upper connector was detached, and the tube tip was submerged in a water container.
Water was then injected by opening valve $V_6$ until the visual cessation of air bubbles from the submerged tip, after which valves $V_6$ and $V_3$ were closed. Finally, the tube tip was reconnected to the air source, and valve $V_5$ was reopened.

%

\subsection{Maximum Fingertip Force}
The maximum fingertip force was measured using a force gauge with a display (DST-1000N, IMADA) in place of the load actuator. With valve $V_1$ open, the finger's base joint was actively moved to press the fingertip against the gauge, while pump pressure was gradually increased and the force recorded.

\subsection{Demonstration of Fingertip Compliance Adjustment}
The experiment was conducted at an initial hydrostatic pressure of 1.2~MPa using the setup shown in Fig.~\ref{fig:task_egg_demonstrations}. 
A raw egg was placed horizontally between the load actuator and fingertip, and the actuator was moved vertically. 
This pressure was experimentally selected to apply a fingertip force as close as practicable to the egg's non-breaking limit.
Sutanto et al. reported a horizontal static failure load of $45.2 \pm 5.52$~N for chicken eggs~\cite{Sutanto2025EggFailure}; thus, 40~N was used as a practical upper target below the reported mean, not as a guaranteed safe load.

\subsection{Hand-Based Task Demonstration}
The qualitative task test was conducted using a three-finger hand (Fig.~\ref{fig:finger_diagram}) and an approximately 50~m-long tube.
The test consisted of picking up a Japanese Industrial Standards (JIS) M8 bolt, gripping an electric hand tool (1.5~kg), and hooking and holding the handle of a toolbox (10~kg). The operator commanded the hand by manually controlling a valve unit in front of the robot.

\section{RESULTS}
To keep the experimental claims traceable to the evaluation protocol, Results
Group~1 supports the 50~m air-condition comparison, Results Group~2 supports
the first 20~m comparison, and Results Group~3 provides the second 20~m cycle.
In each $3\times3$
grid, the columns denote No air, Low air, and High air; Low air and High air
refer to the visually observed 2- and 4~m air-front conditions, respectively.
The rows denote initial pressures of 0.5, 1.0, and 1.5~MPa. Positive and
negative actuator motions are shown separately to expose loading--unloading
loops.

\begin{figure*}[!t]
  \centering
  \includegraphics[width=0.98\textwidth]{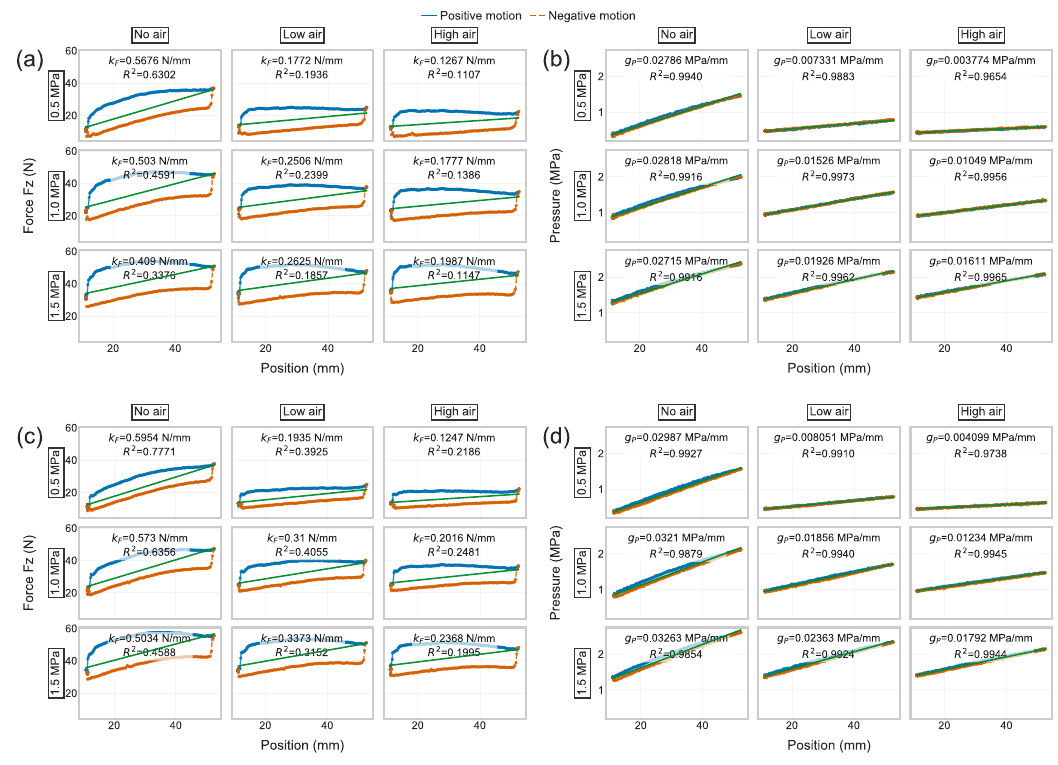}
  \caption{Force--position and pressure--position results: (a) force, 50~m;
  (b) pressure, 50~m (Results Group~1); (c) force, 20~m; and (d) pressure,
  20~m (first cycle, Results Group~2). In each block, columns show No air,
  Low air, and High air; rows show initial pressures of 0.5, 1.0, and 1.5~MPa.
  Low air and High air denote the 2- and 4~m air-front conditions.
  Blue solid and orange dashed traces distinguish positive and negative
  actuator motion (loading and unloading); green lines show full-trajectory
  ordinary least-squares fits with the corresponding slope and $R^2$.}
  \label{fig:group1_results}
  \label{fig:group2_results}
\end{figure*}

\begin{figure}[t]
  \centering
  \includegraphics[width=\columnwidth]{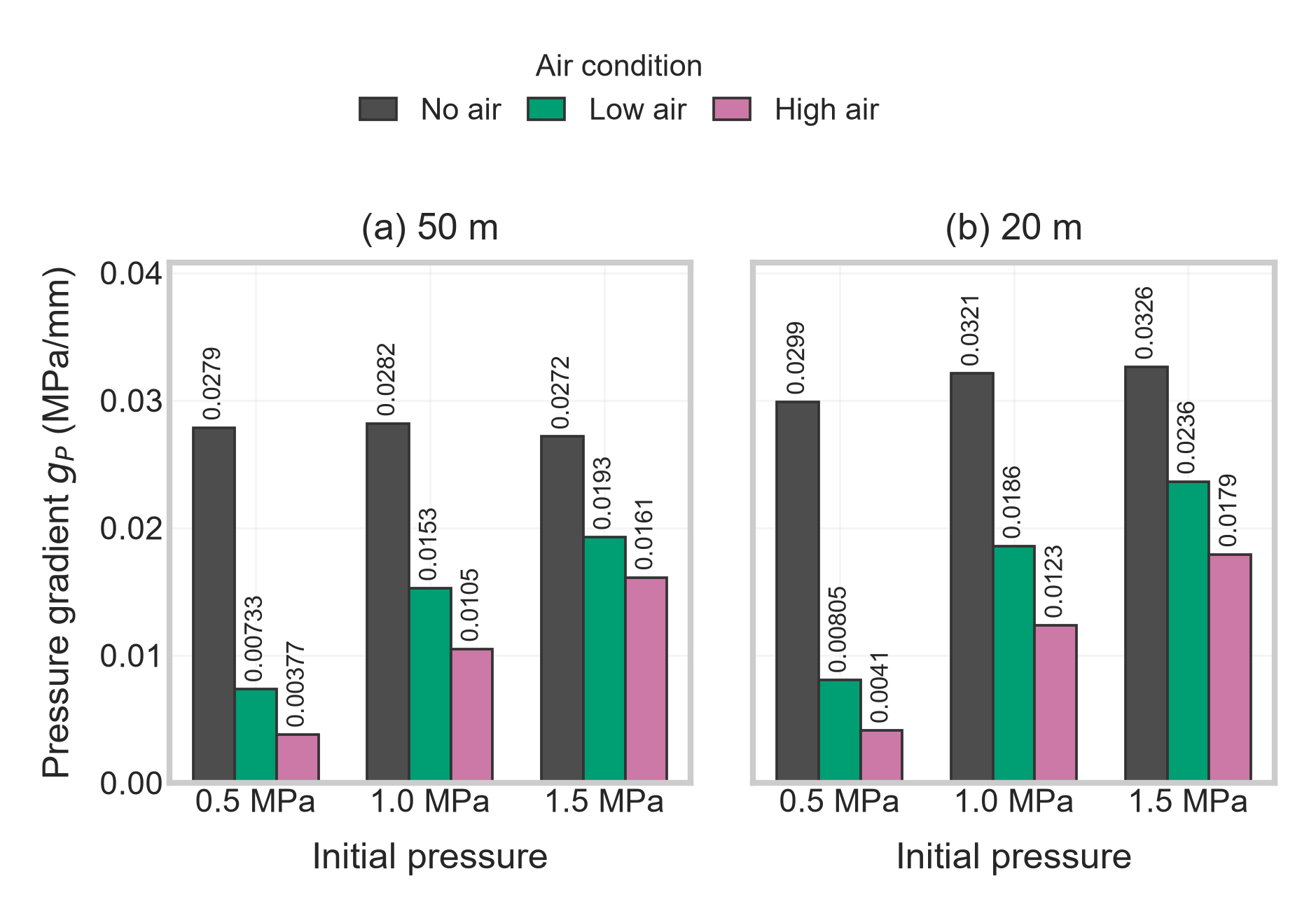}
  \caption{Individual pressure gradients $g_P=dP/dx$ for (a) 50~m and
  (b) 20~m transmission lines. Each panel contains nine bars representing
  three initial pressures and three air conditions; no averaging or error bars
  are applied. Lower $g_P$ corresponds to greater effective pressure-path
  position compliance.}
  \label{fig:pressure_path_summary}
\end{figure}

Figure~\ref{fig:group1_results}(b) and (d) show the
primary pressure--position results. The pressure gradients $g_P$ for Results Group~1 were
0.0272--0.0282, 0.00733--0.0193, and 0.00377--0.0161~MPa/mm for no air, low
air, and high air, respectively; those for Results Group~2 were 0.0299--0.0326,
0.00805--0.0236, and 0.00410--0.0179~MPa/mm. The no-air pressure slope was
nearly independent of initial pressure, whereas the slopes with enclosed air
increased with initial pressure. The pressure fits were generally close to
linear ($R^2=0.965$--0.997 for Results Group~1 and 0.974--0.994 for Results Group~2). In
both transmission configurations, increasing the air condition from No air to Low air and High
air reduced $g_P$ at each initial pressure. Table~\ref{tab:pressure_path_summary}
and Fig.~\ref{fig:pressure_path_summary} summarize this pressure-path result.
The inverse $C_P=1/g_P$ is reported as effective pressure-path position
compliance, not as a material property of the tube alone.
For No air, the configuration with the 50~m main section had a lower mean $g_P$ than that with the 20~m section
(0.0277 versus 0.0315~MPa/mm) and therefore a higher mean inverse compliance
$C_P$ (36.1 versus 31.8~mm/MPa). This length-dependent difference is considered
consistent with a larger contribution from tube expansion in the longer path,
although the present measurements do not isolate tube expansion from water
compressibility and other hydraulic volumes.

The force--position results for Groups~1 and 2 are shown in
Fig.~\ref{fig:group1_results}(a) and (c), respectively. For
Results Group~1, the empirical force-slope ranges $k_F$ were 0.409--0.568,
0.177--0.262, and 0.127--0.199~N/mm for the no-air, low-air, and high-air
conditions, respectively. The corresponding Results Group~2 ranges were
0.503--0.595, 0.193--0.337, and 0.125--0.237~N/mm. The nonlinear force
trajectories show loading--unloading hysteresis, with lower fit $R^2$ values
(0.111--0.630 and 0.199--0.777, respectively). These slopes summarize the
condition-specific measured responses rather than identify fingertip stiffness.

The pressure-gradient summary in Fig.~\ref{fig:pressure_path_summary}
shows each of the nine
$g_P$ values (three initial pressures by three air conditions) separately for
each transmission length, without averaging or error bars.
Table~\ref{tab:pressure_path_summary} provides the complementary aggregate
statistics and the corresponding mean inverse response $C_P$.

Results Group~2 and Results Group~3 use the 20~m main transmission section in successive cycles.
Table~\ref{tab:repeat_compliance} reports the no-air effective pressure-path
compliance $C_P=1/g_P$ together with $R_P^2$, comparing the initial baseline of Group~2 with the baseline of Group~3 after the first air-removal cycle. The final no-air records after the second cycle are not included in this table. The Results Group~3 entries show a return
toward the initial no-air pressure-path response; because these are limited cycle-to-cycle
observations, they are described as a recovery tendency rather than proof of
statistical reversibility.

\begin{table}[!t]
\centering
\caption{Effective pressure-path response for the primary 50 and 20~m comparisons.}
\label{tab:pressure_path_summary}
\scriptsize
\setlength{\tabcolsep}{2.2pt}
\renewcommand{\arraystretch}{1.08}
\resizebox{\columnwidth}{!}{%
\begin{tabular}{@{}llccc@{}}
\toprule
Distance & Air condition & $g_P$ [MPa/mm] & $C_P$ [mm/MPa] & $R_P^2$ range \\
\midrule
\input{plotdata/pressure_path_summary_rows.tex}
\end{tabular}%
}
\parbox{\columnwidth}{\footnotesize\textit{Note:} $g_P$ is the mean $\pm$ sample standard deviation over the three initial pressures, not repeat-trial uncertainty. $C_P$ is the mean of the three individual inverse gradients.}
\end{table}

\begin{table}[!t]
\centering
\caption{Recovery observation of no-air effective pressure-path compliance using
the 20~m transmission line.}
\label{tab:repeat_compliance}
\scriptsize
\setlength{\tabcolsep}{1.6pt}
\renewcommand{\arraystretch}{1.15}
\begin{tabular}{@{}lcc@{}}
\toprule
Initial pressure & Results G2: $C_P$, $R_P^2$ & Results G3: $C_P$, $R_P^2$ \\
\midrule
\input{plotdata/repeat_compliance_rows.tex}
\end{tabular}%
\\[-1mm]
\parbox{\columnwidth}{\footnotesize\textit{Note:} Results Group~2 is the first cycle and
Results Group~3 is the cycle after air removal. Each cell reports
$C_P=1/g_P$ in mm/MPa and the pressure-fit $R_P^2$; bold entries mark the
post-removal Results Group~3 values.}
\end{table}

\begin{figure}[!tb]
  \centering
    \begin{minipage}[t]{1.0\hsize}
        \centering
        \includegraphics[width=\linewidth]{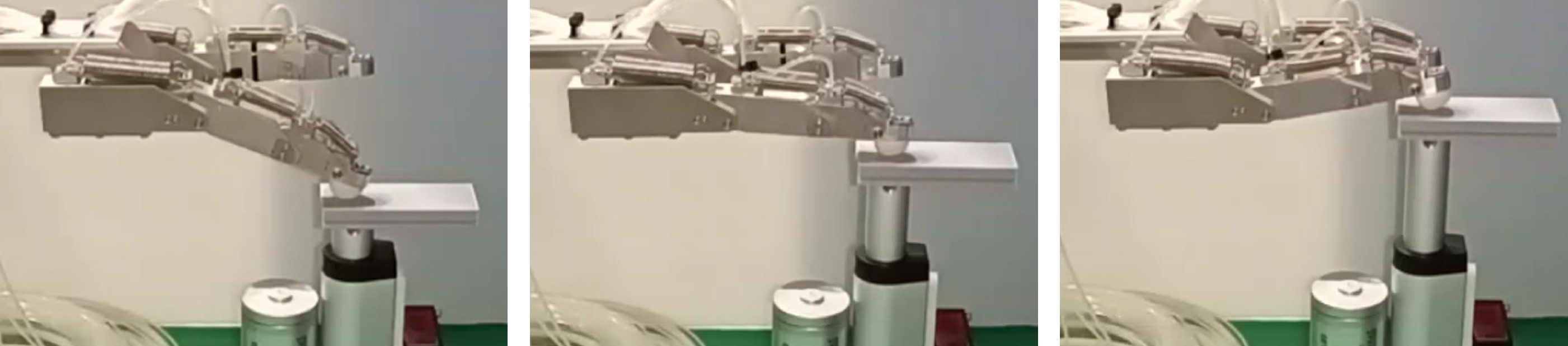}
        \subcaption{Force/pressure-response measurement}
    \end{minipage}
    \\[1.5mm]
    \begin{minipage}[t]{1.0\hsize}
        \centering
        \includegraphics[width=\linewidth]{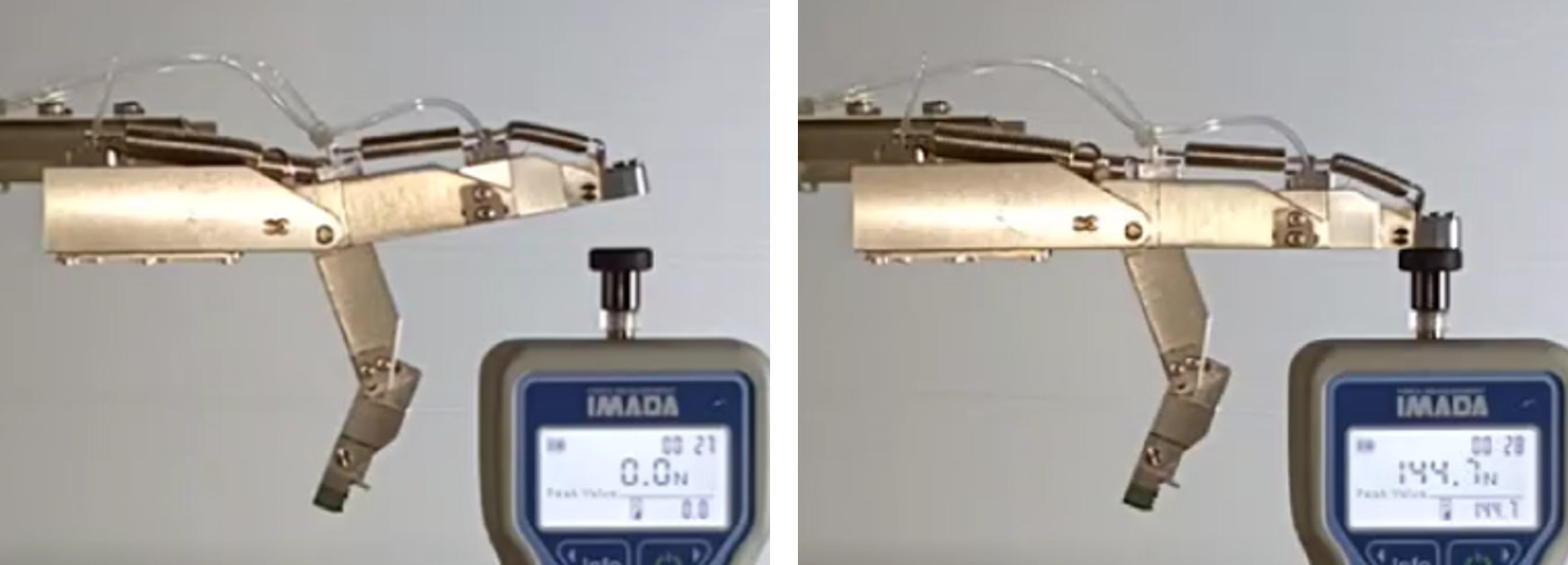}
        \subcaption{Maximum-force measurement}
    \end{minipage}
  \caption{Finger motion and measurement scenes.}
  \label{fig:finger_motion_measurement}
\end{figure}

The measured force was 100~N at a water pressure of 5~MPa, corresponding to
approximately 91\% of the nominal design value of 110~N, and reached 144~N
at a water pressure of 7.5~MPa. The corresponding experimental
scenes are shown in Fig.~\ref{fig:finger_motion_measurement}: (a) shows the
force/pressure-response measurement setup used to obtain the data reported
above, and (b) shows the maximum-fingertip-force measurement.
Maximum-force transmission configuration and repeat-trial variability are not documented; this was a separate test.


\begin{figure}[!tb]
  \centering
    \begin{minipage}[t]{0.36\hsize}
        \centering
        \includegraphics[width=\linewidth]{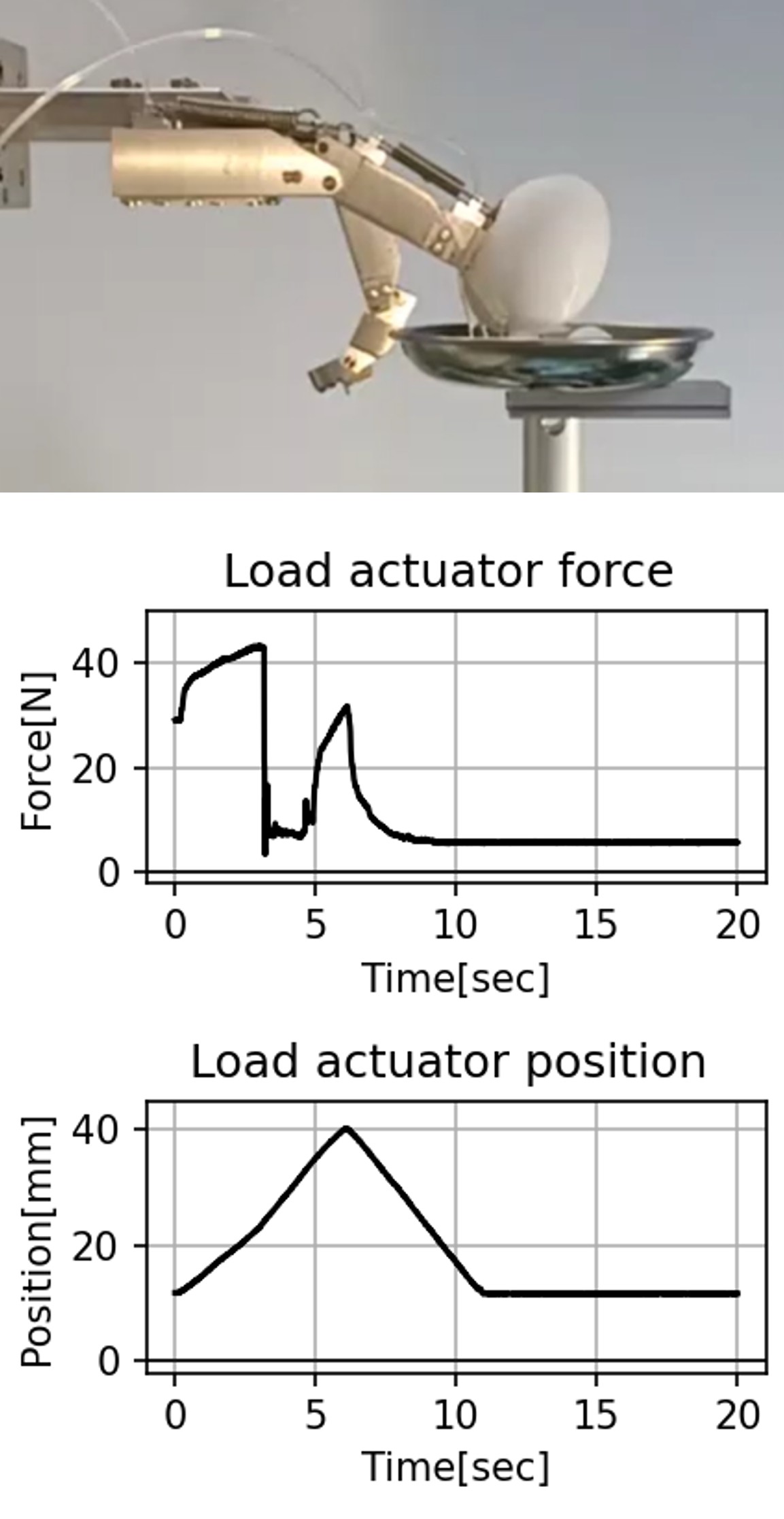}
        \subcaption{No air; observed egg breakage}
    \end{minipage} 
    \begin{minipage}[t]{0.36\hsize}
        \centering
        \includegraphics[width=\linewidth]{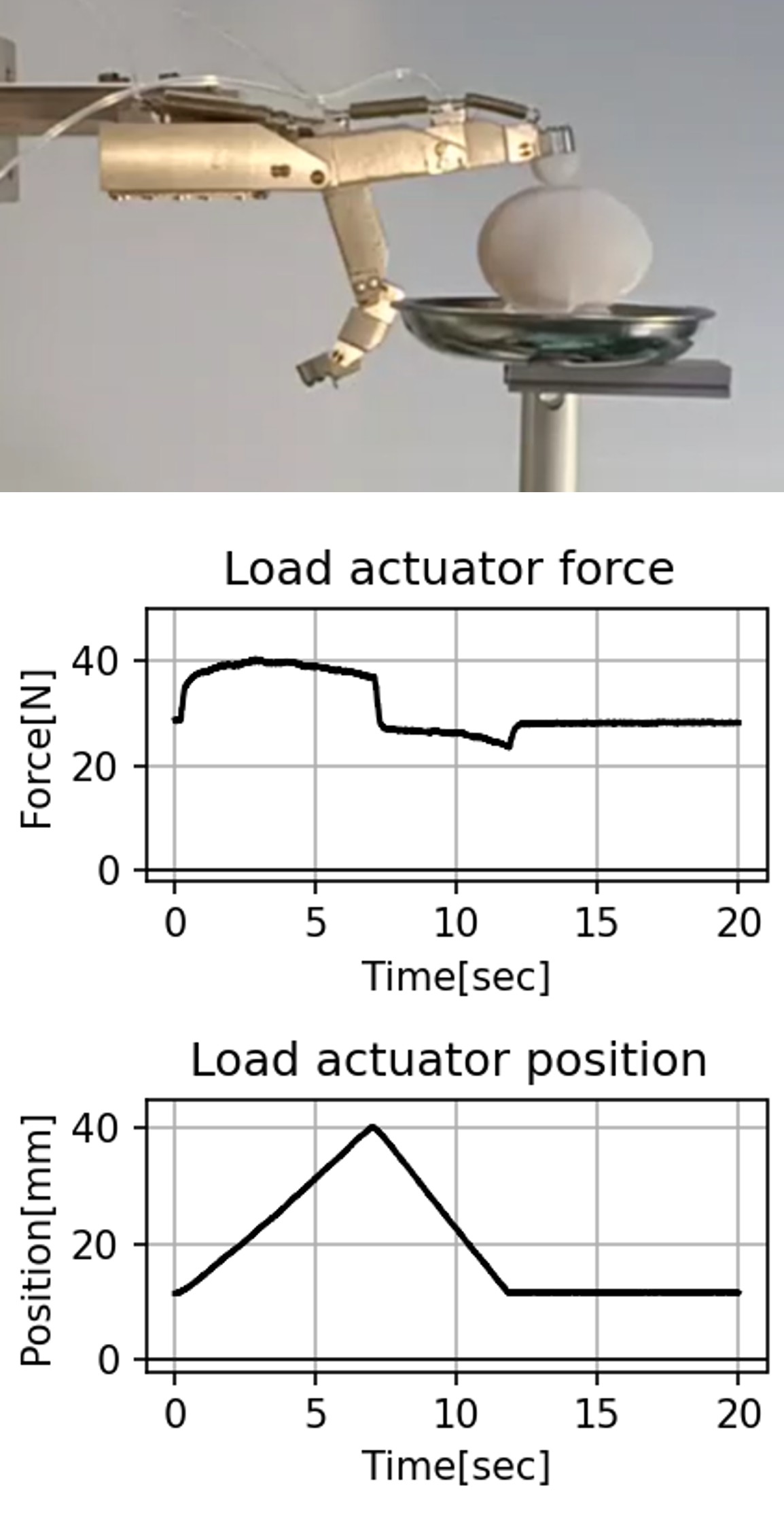}
        \subcaption{High air}
    \end{minipage} \\[1.5mm]
  \caption{Demonstration of fingertip compliance adjustment}
  \label{fig:task_egg_demonstrations}
\end{figure}

Figure~\ref{fig:task_egg_demonstrations} shows the demonstration results for fingertip compliance adjustment. Specifically, Fig.~\ref{fig:task_egg_demonstrations}(a) is the No-air example in which the raw egg broke at around 3~s owing to the increasing force.
In the ``High air'' example in Fig.~\ref{fig:task_egg_demonstrations}(b), the fingertip was observed to maintain contact while accommodating the egg's externally imposed vertical motion, with no observed breakage. This test used the same initial hydrostatic pressure of 1.2~MPa. The illustrated displacement histories differ in timing and are not controlled matched trials; these observations neither quantify contact-force reduction nor establish guaranteed egg protection.

\begin{figure}[!tb]
  \centering
    \begin{minipage}[t]{1.0\hsize}
        \centering
        \includegraphics[width=0.75\linewidth]{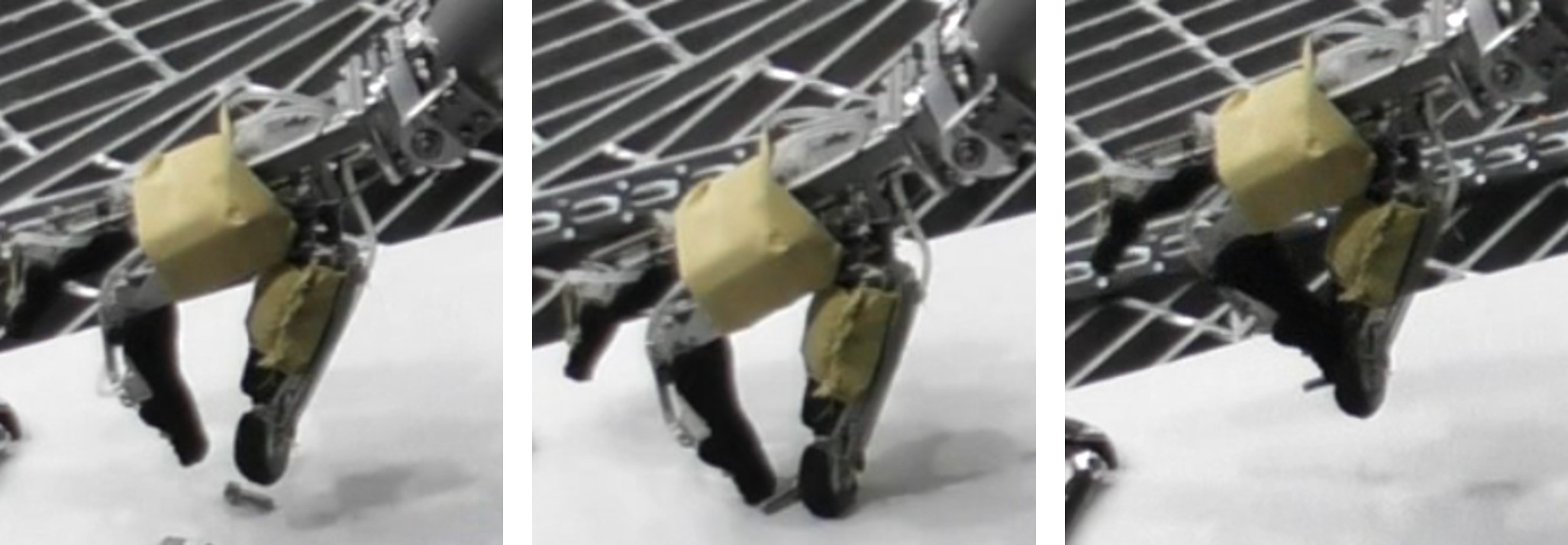}
        \subcaption{Picking up the bolt}
    \end{minipage} \\[1.5mm]
    \begin{minipage}[t]{1.0\hsize}
        \centering
        \includegraphics[width=0.75\linewidth]{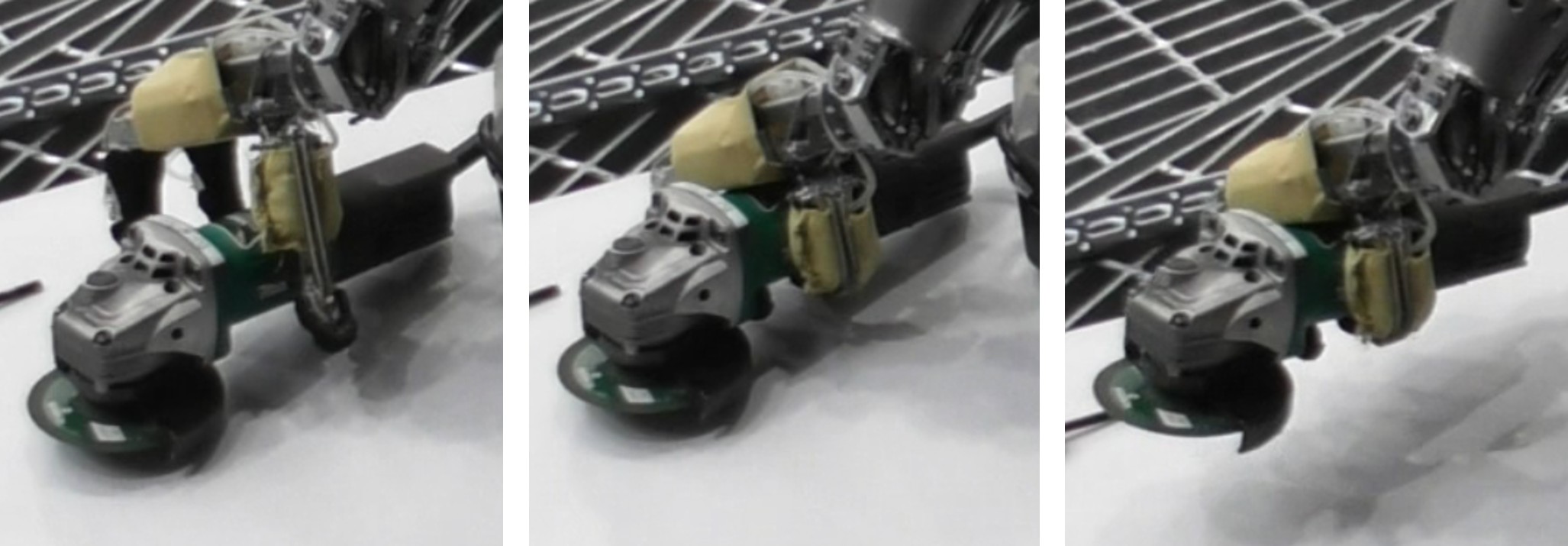}
        \subcaption{Grasping the hand tool}
    \end{minipage} \\[1.5mm]
    \begin{minipage}[t]{1.0\hsize}
        \centering
        \includegraphics[width=0.75\linewidth]{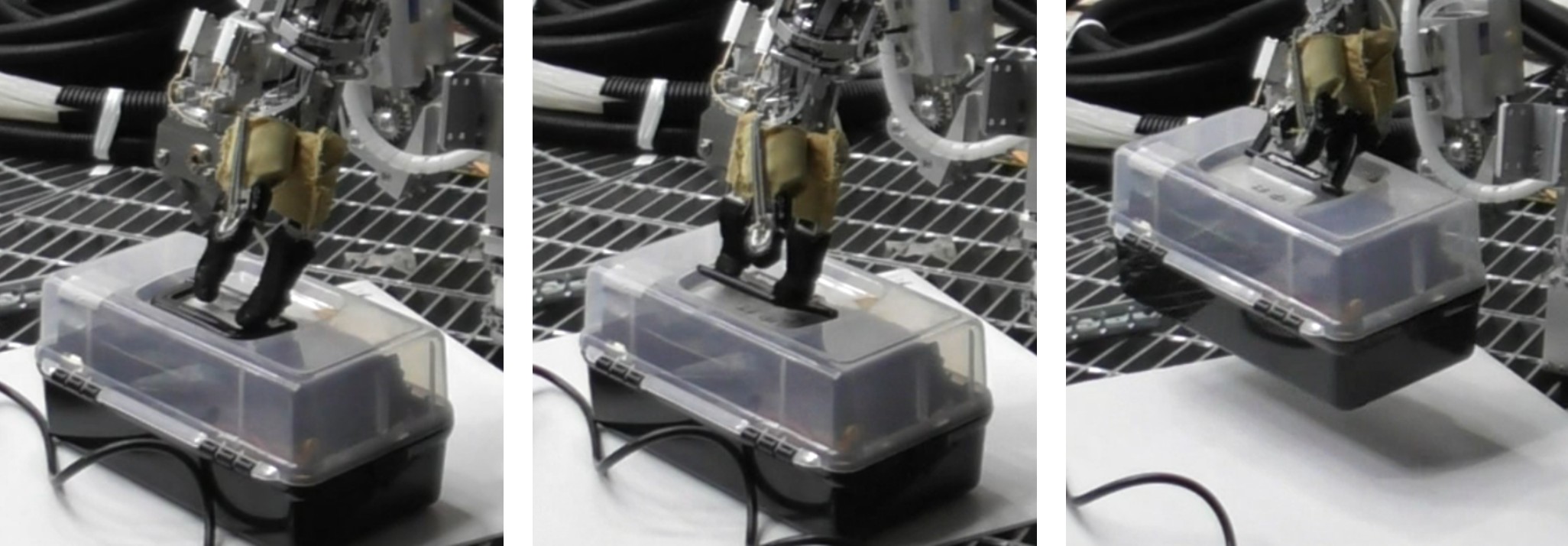}
        \subcaption{Hooking and holding the toolbox handle}
    \end{minipage} \\   
  \caption{Task demonstrations with the developed hand.}
  \label{fig:task_demonstrations}
\end{figure}

During the qualitative three-finger task demonstration through an approximately 50~m tube, the hand picked up a JIS M8 bolt, gripped a hand tool (1.5~kg), and hooked and held the handle of a toolbox (10~kg). During bolt pickup, contact between the fingertip and the ground was tolerated. In particular, when gripping the hand tool, the fingers were qualitatively observed to conform well to the shape of the object owing to their compliance. Furthermore, when hooking the toolbox handle, finger compliance appeared to facilitate insertion of the fingertips beneath the handle. The demonstrations are shown in Fig.~\ref{fig:task_demonstrations}.

\section{DISCUSSION}
The pressure--position relations provide the primary quantitative evidence
that the air section introduced inherent effective compliance into
the 20- and 50~m water-hydraulic transmission paths. For each initial pressure,
$g_P$ decreased from No air to Low air and High air. Under the air conditions,
$g_P$ also increased with initial pressure. These observations are consistent
with pressure-dependent compression of enclosed air, but the present experiment
does not separate air compression from tube expansion, water compressibility,
or dead volume through a predictive pressure--volume model.
The No-air distance comparison points in the same direction: the 50~m path's
lower $g_P$ (equivalently, higher $C_P$) is considered attributable in part to
the greater expansion volume available in the longer tube. This is a physical
interpretation rather than an isolated causal measurement.

Finger-side piston friction and hysteresis limit elastic interpretation of the empirical force slopes, which describe the complete mechanism's response. Pressure gradients provide quantitative support for effective path tunability, rather than identifying individual component properties. The integrated architecture combines qualitative No-air firm grasping with High-air accommodation of externally displaced egg contact; the egg demonstration does not quantify contact-force reduction or safety. Remote pressure generation avoids placing major heat-generating actuators at the hand, but thermal performance, sealing, submersion durability, impact resistance, radiation resistance, and dynamic transmission loss were not evaluated.


Visual air-front setting and the diffuse air--water interface (bubbles mixed with water) limit quantitative evaluation and reproducibility. Future replacement of valve $V_3$ in Fig.~\ref{fig:air_compliance_principle} with a solenoid valve could control opening duration; automation was not implemented here.

Air-mediated adjustment requires no hand-side sensors; hazardous and wet environments motivate the design, but environmental performance is untested.
However, this design may introduce a slower response, hysteresis, temperature sensitivity, and dependence on the distribution of compressed air in the tube.
Future integration with a model of pressure-induced volume loss from tube expansion and air/water compression \cite{nakamura2026shape} could support active control of fingertip posture and flexibility.

\section{CONCLUSION}
This paper presented an electronics-free water-hydraulic hand with opposing fingers for firm grasping and pinching, driven
through small-diameter water-filled tubes with main sections up to 50~m. In both 20 and 50~m single-finger configurations, Low air and High air reduced the pressure gradient relative to No air at every tested initial pressure, quantitatively demonstrating adjustment of effective transmission-path compliance.

The measured maximum fingertip force was 144~N at 7.5~MPa. Under No air, the three-finger hand performed qualitative bolt, tool, and toolbox-handle tasks through a 50~m tube. Upstream air addition and removal tune the effective path response, and under High air a single finger maintained contact while accommodating an externally displaced egg without observed breakage. These results bring firm grasping and compliant contact together within the same architecture. The force slopes summarize empirical fingertip responses, while pressure gradients characterize path tunability. Manual air removal produced a return tendency toward the no-air response; repeated trials and a pressure--volume model remain necessary to establish statistical reversibility and attribute the compliance to individual physical components.

\bibliographystyle{IEEEtran}
\bibliography{Nakamura_etal_Humanoids2026}

\end{document}

%% file: plotdata/pressure_path_summary_rows.tex
50 m & No air & $0.0277 \pm 0.0005$ & $36.1$ & $0.992$--$0.994$ \\
50 m & Low air & $0.0140 \pm 0.0061$ & $84.6$ & $0.988$--$0.997$ \\
50 m & High air & $0.0101 \pm 0.0062$ & $140.8$ & $0.965$--$0.997$ \\
\midrule
20 m & No air & $0.0315 \pm 0.0015$ & $31.8$ & $0.985$--$0.993$ \\
20 m & Low air & $0.0167 \pm 0.0079$ & $73.5$ & $0.991$--$0.994$ \\
20 m & High air & $0.0115 \pm 0.0070$ & $126.9$ & $0.974$--$0.994$ \\
\bottomrule

%% file: plotdata/repeat_compliance_rows.tex
0.5 MPa & $33.5$, $0.993$ & $\mathbf{31.9}$, $\mathbf{0.989}$ \\
1.0 MPa & $31.2$, $0.988$ & $\mathbf{30.4}$, $\mathbf{0.988}$ \\
1.5 MPa & $30.7$, $0.985$ & $\mathbf{30.2}$, $\mathbf{0.984}$ \\
\bottomrule